\documentclass[letterpaper, 10 pt, conference]{ieeeconf}

\IEEEoverridecommandlockouts
\usepackage{cite}
\usepackage{amsmath,amssymb,amsfonts}
\usepackage{graphicx}
\usepackage{textcomp}
\usepackage{xcolor}
\usepackage{booktabs}
\usepackage{multirow}
\usepackage{url}
\usepackage{etoolbox}

\makeatletter
\patchcmd{\@maketitle}{\vskip0.25in{\LARGE\@title\par}\vskip1.0em\par}%
  {\vskip0.25in{\LARGE\@title\par}\vskip0.52em\par}%
  {\typeout{** root_arxiv: patched title-to-author spacing **}}{\typeout{** root_arxiv: WARNING patch failed; title spacing unchanged **}}%
\def\@IEEEauthorblockconfadjspace{-0.38em}%
\makeatother

\graphicspath{{Figures/}}

\title{\LARGE \bf
Edge AI for Automotive Vulnerable Road User Safety: Deployable Detection via Knowledge Distillation
}

\author{Akshay Karjol,~\IEEEmembership{Senior Member, IEEE,}
and Darrin M. Hanna,~\IEEEmembership{Member, IEEE}
\thanks{A. Karjol and D. M. Hanna are with the Department of Electrical and Computer Engineering, Oakland University, Rochester, MI 48309 USA (e-mail: akshaykarjol@oakland.edu; hanna@oakland.edu).}
\thanks{Manuscript received April 2026.}}

\begin{document}

\IEEEaftertitletext{%
  \vspace{-1.2\baselineskip}%
  \begingroup\setlength{\fboxsep}{2pt}\setlength{\fboxrule}{0.4pt}%
  \noindent
  \fbox{\begin{minipage}{\dimexpr\textwidth-2\fboxsep-2\fboxrule\relax}\footnotesize\noindent
  This work has been submitted to the IEEE for possible publication. Copyright may be transferred without notice, after which this version may no longer be accessible.%
  \end{minipage}}%
  \endgroup\vspace{-0.18\baselineskip}%
}

\maketitle
\thispagestyle{empty}
\pagestyle{empty}

\begin{abstract}

Deploying accurate object detection for Vulnerable Road User (VRU) safety on edge hardware requires balancing model capacity against computational constraints. Large models achieve high accuracy but fail under INT8 quantization required for edge deployment, while small models sacrifice detection performance. This paper presents a knowledge distillation (KD) framework that trains a compact YOLOv8-S student (11.2M parameters) to mimic a YOLOv8-L teacher (43.7M parameters), achieving 3.9$\times$ compression while preserving quantization robustness. We evaluate on full-scale BDD100K (70K training images) with Post-Training Quantization to INT8. The teacher suffers catastrophic degradation under INT8 ($-$23\% mAP), while the KD student retains accuracy ($-$5.6\% mAP). Analysis reveals that KD transfers precision calibration rather than raw detection capacity: the KD student achieves 0.748 precision versus 0.653 for direct training at INT8, a 14.5\% gain at equivalent recall, reducing false alarms by 44\% versus the collapsed teacher. At INT8, the KD student exceeds the teacher's FP32 precision (0.748 vs.\ 0.718) in a model 3.9$\times$ smaller. These findings establish knowledge distillation as a requirement for deploying accurate, safety-critical VRU detection on edge hardware.

\end{abstract}


\section{Introduction}

Road traffic crashes remain a leading cause of preventable death worldwide. The World Health Organization reports 1.19 million annual road traffic deaths, with road injuries ranking as the leading cause of death for ages 5 to 29 \cite{who2023road}. Vulnerable Road Users (VRUs), defined as pedestrians, cyclists, and motorcyclists, account for over half of all traffic fatalities globally \cite{who2023road}. In the United States, 7,314 pedestrians and 1,166 cyclists died in 2023 \cite{nhtsa2024pedestrian, nhtsa2024cyclist}.

Automated driving systems from SAE Level 2 to Level 4/5 rely on real-time perception to detect VRUs \cite{chen2021dnn, yue2020effectiveness}. Deep learning detectors, particularly the YOLO (You Only Look Once) family, achieve strong performance on driving datasets \cite{redmon2016yolo, jocher2023yolov8}. However, deploying these models on edge hardware presents challenges. Edge AI accelerators such as NVIDIA Jetson impose strict constraints on model size, memory, and latency \cite{liu2019edge, reuther2019survey}. A 200 ms perception delay at 60 mph equals 5.3 m of uncontrolled travel \cite{liu2019edge}. INT8 quantization reduces memory by 4$\times$ and enables efficient integer arithmetic, making it the standard precision for edge deployment \cite{jacob2018quantization, wu2020integer}.

INT8 quantization, however, creates a dilemma for VRU safety. Large models such as YOLOv8-L (43.7M parameters) achieve high accuracy but degrade under INT8 quantization. Prior studies report 3 to 7\% mAP loss for YOLOv5-S and YOLOv8-S under TensorRT INT8 \cite{wu2020integer, chen2025int8}. Larger models with wider weight distributions compress poorly to 8-bit representation \cite{nagel2021white}. Small models such as YOLOv8-S (11.2M parameters) meet computational constraints but sacrifice performance on underrepresented classes \cite{oksuz2024imbalance}. Neither approach alone satisfies the requirements for deployable, accurate VRU detection.

Knowledge distillation (KD) trains a compact student to match the soft outputs of a larger teacher \cite{hinton2015distilling}. Temperature scaling softens the output distribution, encoding class similarity absent from one-hot labels \cite{hinton2015distilling}. Feature-based methods extend this by aligning intermediate representations between student and teacher \cite{romero2015fitnets, yim2017gift}, but require additional loss terms and increase training cost \cite{gou2021knowledge}. FGD applies separate distillation losses to foreground and background regions \cite{yang2022fgd}. PKD uses correlation-based alignment for heterogeneous architectures \cite{cao2022pkd}. Vanilla KD achieves equivalent accuracy at scale without additional loss terms \cite{yang2023vanillakd}.

Prior work shows that vanilla KD transfers detection accuracy at scale \cite{yang2023vanillakd}. Whether KD also transfers confidence calibration (precision versus recall) and confers INT8 quantization robustness has not been studied. Answering both determines whether a KD-trained student can replace a large teacher in edge deployment.

This paper addresses this gap through systematic evaluation on BDD100K \cite{yu2020bdd100k}, the largest driving dataset available. We train YOLOv8-S students with and without teacher guidance, then quantize all models to INT8 using TensorRT Post-Training Quantization (PTQ) \cite{nvidia2024tensorrt}. Our experiments show that KD transfers confidence calibration rather than raw detection capacity. This calibration survives INT8 quantization. The KD student achieves 44\% fewer false alarms than the collapsed teacher at INT8. This improvement matters for automated driving where false alarms degrade driver trust \cite{lee2004trust}.

The main contributions of this work are:
\begin{itemize}
\item Empirical demonstration that KD-trained models degrade significantly less under INT8 PTQ ($-$5.6\% mAP) compared to large teachers ($-$23\% mAP), establishing KD as essential for edge-deployable VRU detection.
\item Analysis showing KD transfers precision calibration rather than recall: KD student achieves +14.5\% precision gain at INT8 with equivalent recall, reducing false alarms by 44\% versus the collapsed teacher, critical for edge-deployed automotive safety systems.
\item Full-scale validation on BDD100K (70K training images) with VRU-specific evaluation across pedestrian, cyclist, and motorcyclist classes.
\end{itemize}

\section{Related Work}

\subsection{VRU Detection for Autonomous Driving}

Detecting VRUs (pedestrians, cyclists, motorcyclists) is a core challenge for autonomous driving safety \cite{chen2021dnn, yue2020effectiveness}. Early approaches used hand-crafted features such as Histogram of Oriented Gradients (HOG) with Support Vector Machines \cite{dalal2005hog}. Deep learning transformed the field. Convolutional neural networks achieve higher accuracy than classical methods \cite{chen2021dnn}.

The YOLO (You Only Look Once) family dominates real-time detection due to its single-stage architecture \cite{redmon2016yolo, wang2023yolov7}. YOLOv8 introduced anchor-free detection, the C2f feature fusion module, and spatial attention \cite{jocher2023yolov8}.

\subsection{Knowledge Distillation for Object Detection}

Knowledge distillation trains a compact student to match soft outputs from a larger teacher \cite{hinton2015distilling}. Temperature scaling softens probability distributions, encoding class similarity absent from one-hot labels \cite{hinton2015distilling}.

Feature-based distillation aligns intermediate representations between student and teacher \cite{romero2015fitnets, yim2017gift}. FitNets \cite{romero2015fitnets} introduced hint layers to guide student features toward teacher features. For object detection, FGD \cite{yang2022fgd} separates distillation losses for foreground and background regions. PKD uses correlation-based alignment for heterogeneous architectures, achieving 4.1 to 4.8\% mAP improvement on COCO \cite{cao2022pkd}.

Recent work applies KD to YOLO architectures. LKD-YOLOv8 \cite{liu2024lkd} combines masked generative distillation with lightweight convolution. CrossKD \cite{wang2024crosskd} introduces cross-head distillation to relieve contradictory supervision. Gou et al.\ \cite{gou2021knowledge} categorize KD methods into response-based, feature-based, and relation-based approaches. While advanced methods achieve incremental gains, vanilla KD remains effective at scale \cite{yang2023vanillakd}. The interaction between KD and quantization robustness remains unexplored.

\subsection{Model Quantization for Edge Deployment}

Quantization reduces model precision from 32-bit floating point to lower bit-widths \cite{jacob2018quantization, wu2020integer}. INT8 quantization offers 4$\times$ memory reduction and efficient integer arithmetic on hardware accelerators \cite{jacob2018quantization}.

Post-Training Quantization (PTQ) converts pretrained models without retraining \cite{nagel2021white}. PTQ uses a calibration dataset to determine scaling factors \cite{jacob2018quantization, nagel2021white}. PTQ requires minutes rather than days but can degrade accuracy, particularly for large models with wide weight distributions \cite{wu2020integer}. Quantization-Aware Training (QAT) maintains higher accuracy at greater computational cost \cite{esser2020lsq}.

NVIDIA TensorRT provides optimized inference with PTQ support \cite{nvidia2024tensorrt}.

\subsection{Confidence Calibration in Neural Networks}

Modern neural networks produce overconfident predictions \cite{guo2017calibration}. Temperature scaling learns a single parameter to scale logits, improving calibration without affecting accuracy \cite{guo2017calibration}. Whether KD training with temperature-scaled soft labels transfers calibration to the student, and whether this calibration survives INT8 quantization, has not been studied.

\subsection{False Alarm Impact on ADAS and AV Acceptance}

False alarms impact ADAS and AV effectiveness and acceptance \cite{lee2004trust}. Excessive false warnings erode driver trust and lead to system deactivation \cite{khastgir2020watchman}. For VRU detection, false alarms are particularly damaging given VRU rarity. Precision is critical for deployable systems.

\section{Methodology}

To address both open questions, we design a controlled experimental framework comparing three training configurations: a large teacher, a compact direct student, and a compact KD student. All three are evaluated at FP32 and INT8. We formulate the detection problem, define the KD objective, and detail the INT8 quantization procedure.

\subsection{Problem Formulation}

Let $f_T: \mathcal{X} \rightarrow \mathcal{Y}$ denote a large teacher detector and $f_S: \mathcal{X} \rightarrow \mathcal{Y}$ denote a compact student detector, where $\mathcal{X}$ represents input images and $\mathcal{Y}$ represents detection outputs (bounding boxes and class probabilities). The task is VRU detection in driving scenes: pedestrians, cyclists, and motorcyclists across the BDD100K dataset. The objective is to train $f_S$ such that quantized inference $Q(f_S)$ preserves accuracy while meeting edge deployment constraints.

We compare three training strategies:
\begin{itemize}
\item Teacher: YOLOv8-L trained directly on ground truth labels
\item Direct Student: YOLOv8-S trained directly on ground truth labels
\item KD Student: YOLOv8-S trained with teacher guidance
\end{itemize}

This comparison isolates the effect of knowledge distillation: the direct and KD students share identical architectures, differing only in training supervision. By evaluating all three models at FP32 and INT8, we can determine whether KD-trained models exhibit superior quantization robustness.

\subsection{Knowledge Distillation Framework}

We adopt a task-dampened formulation where the KD training objective combines task loss with distillation losses:
\begin{equation}
\mathcal{L}_{\text{total}} = \alpha \cdot \mathcal{L}_{\text{task}} + \beta \cdot \mathcal{L}_{\text{logit}} + \gamma \cdot \mathcal{L}_{\text{feat}}
\end{equation}
where $\mathcal{L}_{\text{task}}$ is the standard detection loss (classification + localization), $\mathcal{L}_{\text{logit}}$ is the logit-based distillation loss, $\mathcal{L}_{\text{feat}}$ is the feature-based distillation loss, and $\alpha$, $\beta$, $\gamma \in [0,1]$ are scalar weights controlling the contribution of each loss term.

The logit distillation loss uses KL divergence with temperature scaling:
\begin{equation}
\mathcal{L}_{\text{logit}} = T^2 \cdot \text{KL}\left(\sigma\left(\frac{\mathbf{z}_S}{T}\right) \| \sigma\left(\frac{\mathbf{z}_T}{T}\right)\right)
\end{equation}
where $\mathbf{z}_S$ and $\mathbf{z}_T$ are student and teacher logits, $\sigma$ is softmax, $T$ is temperature, and $\text{KL}$ denotes the Kullback-Leibler divergence. The $T^2$ factor compensates for gradient magnitude reduction at higher temperatures. Higher $T$ produces softer distributions, encoding class similarity absent from one-hot labels.

The feature distillation loss aligns intermediate representations using L2 distance:
\begin{equation}
\mathcal{L}_{\text{feat}} = \|\mathbf{F}_S - \mathbf{F}_T\|_2^2
\end{equation}
where $\mathbf{F}_S$ and $\mathbf{F}_T$ are student and teacher feature maps from the backbone.

The teacher is frozen during KD training. Only the student parameters update. We set $\alpha = 0.5$, $\beta = 0.3$, $\gamma = 0.02$, and $T = 10$. The task loss weight $\alpha = 0.5$ down-weights ground-truth supervision relative to teacher guidance, increasing the influence of soft distributions on student training. We validate this choice through ablation in Section~\ref{sec:ablation}. The low feature loss weight ($\gamma = 0.02$) limits the constraint on student intermediate representations.

\subsection{INT8 Quantization}

We apply Post-Training Quantization (PTQ) to convert FP32 models to INT8 \cite{jacob2018quantization}. PTQ determines scaling factors by running forward passes on a calibration dataset. We use TensorRT for quantization and inference. Calibration images are sampled from an unlabeled split separate from training and evaluation data.

\section{Experimental Setup}

\subsection{Dataset}

\begin{figure}[htbp]
\centering
\includegraphics[width=\columnwidth]{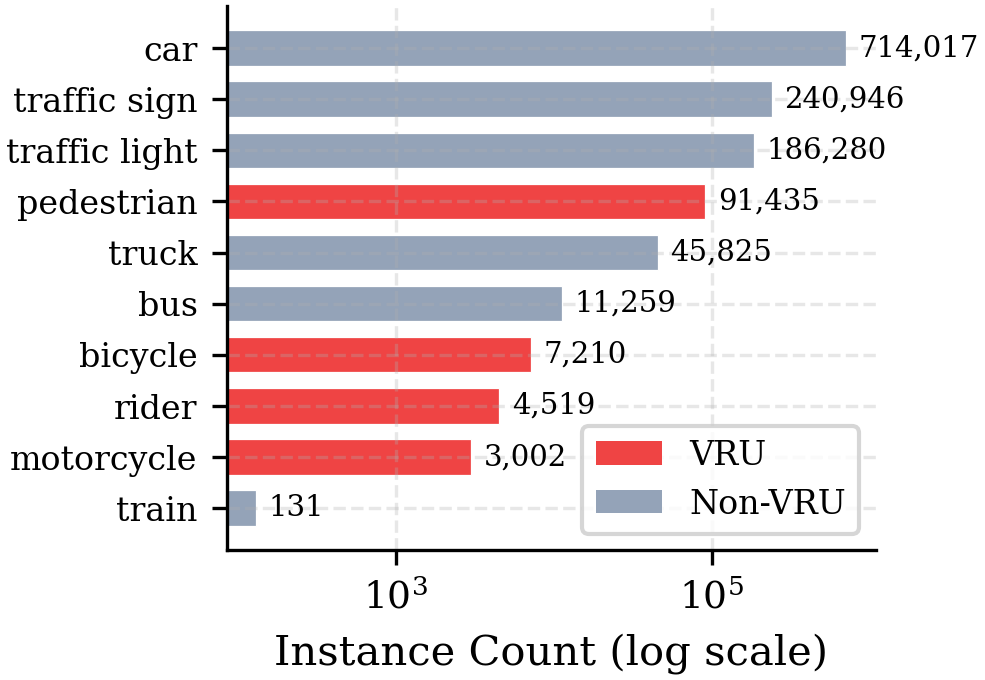}
\caption{BDD100K class distribution showing severe imbalance. VRU classes (red) are significantly underrepresented compared to dominant classes such as cars.}
\label{fig:class_distribution}
\end{figure}

We evaluate on BDD100K \cite{yu2020bdd100k}, which contains 100K driving images annotated with 10 object classes. We follow the standard split: 70K training, 10K validation, and 20K test images. Test set annotations are not publicly available, so all accuracy evaluations use the validation set. For INT8 calibration, we sample 1,024 images from the test split with a fixed random seed to prevent data leakage. The dataset exhibits severe class imbalance: cars dominate with 714K instances, while VRU classes are underrepresented (pedestrians: 91K, riders: 5K, bicycles: 7K, motorcycles: 3K), as shown in Fig.~\ref{fig:class_distribution}.

We define three VRU groups: pedestrians, cyclists (rider+bicycle), and motorcyclists (rider+motorcycle). BDD100K uses overlapping bounding boxes for cyclists and motorcyclists, which we evaluate as combined groups.

\subsection{Model Architecture}

We select YOLOv8 for its mature TensorRT INT8 pipeline and deployment ecosystem \cite{jocher2023yolov8}. Newer variants (YOLOv9, YOLOv10, YOLO11) offer 1 to 2\% mAP improvements \cite{jha2025yolo}, but YOLOv8 remains fastest for edge deployment \cite{zhang2025yolospeed}. Generalization of KD-induced quantization robustness to other architectures remains to be validated.

We use YOLOv8-L (43.7M parameters) as the teacher and YOLOv8-S (11.2M parameters) as the student, achieving 3.9$\times$ parameter compression. Both models use identical input resolution (640$\times$640) and are initialized from COCO-pretrained weights.

\subsection{Training Configuration}

All models train for 50 epochs using AdamW with cosine annealing and standard YOLOv8 augmentation (mosaic, mixup, HSV jitter). The teacher uses batch size 64 and learning rate 0.00283. Both students use batch size 256 and learning rate 0.004. All models use 3-epoch warmup and early stopping patience of 30 epochs. The teacher trains first and remains frozen during student training.

Training uses cloud GPU infrastructure (NVIDIA RTX PRO 6000 Blackwell, 96GB VRAM). The teacher requires 7 hours, the direct student 2.7 hours, and the KD student 4.9 hours. The KD overhead reflects teacher forward passes during distillation.

\subsection{Evaluation Metrics}

We report standard detection metrics: mAP50 (mean Average Precision at IoU 0.5), mAP50-95 (averaged over IoU thresholds 0.5--0.95), precision, and recall. For safety-critical analysis, we compute False Alarm Rate (FAR = 1 $-$ Precision) to quantify incorrect positive predictions that can erode driver trust \cite{lee2004trust}. We evaluate VRU-specific mAP50 across the three VRU groups (pedestrians, cyclists, motorcyclists) and measure inference throughput in frames per second (FPS).

\subsection{Quantization and Deployment}

We quantize models using NVIDIA TensorRT on an RTX 5070 GPU (8GB VRAM). The RTX 5070 is closer to edge deployment conditions than the training hardware. Throughput on the RTX PRO 6000 exceeds 1,500 FPS, which is not representative of deployment conditions. Deployment on automotive-grade hardware such as NVIDIA Jetson remains future work. We evaluate three precision levels: FP32 (baseline), FP16 (half precision), and INT8 (8-bit integer). Throughput is measured as frames per second (FPS) averaged over 1,000 inference iterations.

\section{Results}
\renewcommand{\arraystretch}{1.0}
\setlength{\textfloatsep}{6pt plus 2pt minus 2pt}
\setlength{\floatsep}{6pt plus 2pt minus 2pt}
\setlength{\intextsep}{6pt plus 2pt minus 2pt}

We evaluate all three models (teacher, direct student, KD student) at FP32 and INT8. We establish FP32 performance, then analyze VRU-specific detection, quantization degradation, precision at INT8, throughput, and task loss weight ablation.

\subsection{FP32 Baseline Performance}

\begin{table}[htbp]
\caption{FP32 Detection Performance on BDD100K}
\label{tab:fp32_results}
\centering
\setlength{\tabcolsep}{3pt}
\begin{tabular}{llccccc}
\toprule
\textbf{Model} & \textbf{Params} & \textbf{mAP50} & \textbf{mAP50-95} & \textbf{Precision} & \textbf{Recall} & \textbf{FAR} \\
\midrule
Teacher (L) & 43.7M & 0.595 & 0.345 & 0.718 & 0.554 & 0.282 \\
Direct (S) & 11.2M & 0.533 & 0.300 & 0.633 & 0.480 & 0.367 \\
KD (S) & 11.2M & 0.532 & 0.300 & 0.716 & 0.483 & 0.284 \\
\bottomrule
\end{tabular}
\end{table}

Table~\ref{tab:fp32_results} presents FP32 detection performance on the BDD100K validation set. The teacher achieves 0.595 mAP50, outperforming both students (0.532 to 0.533 mAP50). At FP32, KD and direct students achieve identical mAP50. The KD student achieves higher precision (0.716 vs. 0.633) at equivalent recall (0.483 vs. 0.480). The KD student matches the teacher's FAR (0.284 vs. 0.282), while the direct student has 30\% higher FAR (0.367). KD transfers confidence calibration from teacher to student.

\subsection{VRU-Specific Performance}

\begin{table}[htbp]
\caption{VRU Group Detection Performance (mAP50, FP32)}
\label{tab:vru_results}
\centering
\begin{tabular}{lcccc}
\toprule
\textbf{Model} & \textbf{Pedestrian} & \textbf{Cyclist} & \textbf{Motorcyclist} & \textbf{VRU Avg} \\
\midrule
Teacher (L) & 0.685 & 0.522 & 0.503 & 0.559 \\
Direct (S) & 0.613 & 0.453 & 0.440 & 0.492 \\
KD (S) & 0.613 & 0.450 & 0.440 & 0.491 \\
\bottomrule
\end{tabular}
\end{table}

Table~\ref{tab:vru_results} presents VRU group detection performance. The teacher leads by 6.3 to 7.2 mAP50 points across all VRU groups. KD and direct students achieve equivalent VRU performance. KD does not boost recall on rare classes. The gap is architectural: 11.2M parameters cannot match 43.7M capacity for underrepresented classes.

\subsection{Quantization Impact}

\begin{table}[htbp]
\caption{Quantization Impact: FP32 to INT8}
\label{tab:quant_results}
\centering
\begin{tabular}{lcccc}
\toprule
\textbf{Model} & \textbf{FP32} & \textbf{INT8} & \textbf{$\Delta$mAP} & \textbf{$\Delta$\%} \\
\midrule
Teacher (L) & 0.595 & 0.456 & $-$0.139 & $-$23.4\% \\
Direct (S) & 0.533 & 0.502 & $-$0.031 & $-$5.8\% \\
KD (S) & 0.532 & 0.502 & $-$0.030 & $-$5.6\% \\
\bottomrule
\end{tabular}
\end{table}

\begin{figure}[htbp]
\centering
\includegraphics[width=\columnwidth]{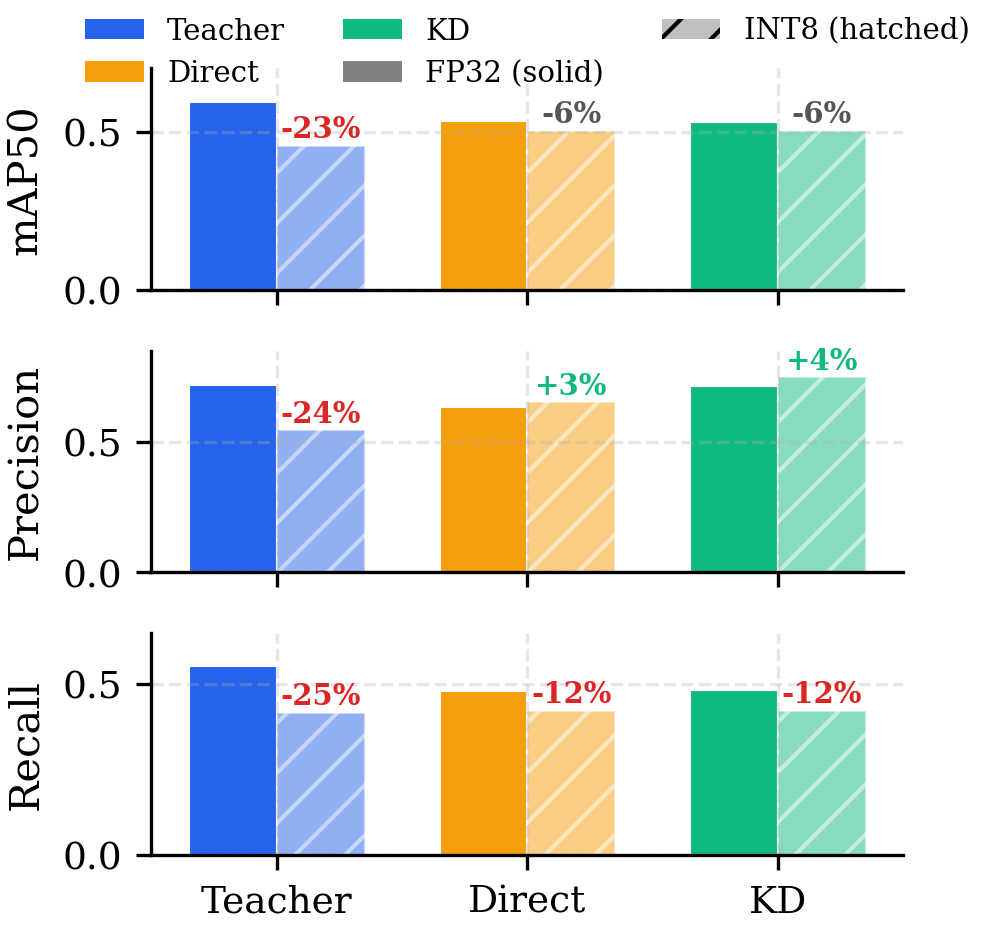}
\caption{Quantization impact: FP32 (solid) vs INT8 (hatched). Teacher collapses ($-$23\% mAP, $-$24\% precision), students retain performance. KD gains +4\% precision under INT8.}
\label{fig:quantization}
\end{figure}

Table~\ref{tab:quant_results} presents the impact of INT8 quantization. The teacher degrades catastrophically under INT8 ($-$23\% mAP), making it unsuitable for edge deployment despite superior FP32 performance. Fig.~\ref{fig:quantization} shows this comparison. Both students exhibit acceptable degradation ($-$5.6\% to $-$5.8\%). The KD student achieves slightly better retention. Large models are more sensitive to INT8 due to wider weight distributions.

\subsection{Precision Analysis at INT8}

\begin{table}[htbp]
\caption{Precision and False Alarm Rate at INT8}
\label{tab:precision_int8}
\centering
\setlength{\tabcolsep}{3pt}
\begin{tabular}{lcccccc}
\toprule
\textbf{Model} & \textbf{mAP50} & \textbf{Precision} & \textbf{$\Delta$Prec} & \textbf{Recall} & \textbf{FAR} & \textbf{$\Delta$FAR} \\
\midrule
Teacher (L) & 0.456 & 0.546 & baseline & 0.416 & 0.454 & baseline \\
Direct (S) & 0.502 & 0.653 & $+$20\% & 0.421 & 0.347 & $-$24\% \\
KD (S) & 0.502 & 0.748 & $+$37\% & 0.423 & 0.252 & $-$44\% \\
\bottomrule
\end{tabular}
\end{table}

\begin{figure}[htbp]
\centering
\includegraphics[width=\columnwidth]{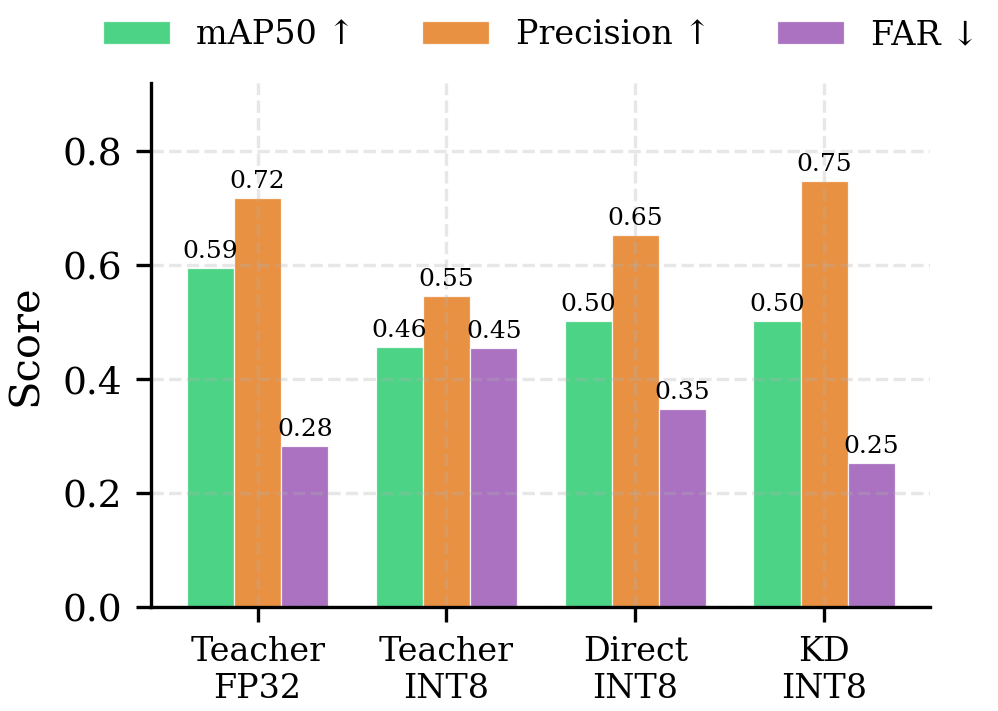}
\caption{Deployment comparison: KD INT8 achieves highest precision (0.75) and lowest FAR (0.25) among all deployable configurations.}
\label{fig:deployment}
\end{figure}

Table~\ref{tab:precision_int8} compares precision, recall, and false alarm rates at INT8, using the collapsed teacher as baseline. The KD student achieves 0.748 precision ($+$37\% vs. teacher) and 0.252 FAR ($-$44\%), exceeding the teacher's FP32 precision (0.718) in a model 3.9$\times$ smaller. The direct student achieves $+$20\% precision and $-$24\% FAR. All three models produce equivalent recall (0.416 to 0.423), confirming that KD transfers precision calibration rather than detection capacity. Fig.~\ref{fig:deployment} summarizes the deployment comparison.

\subsection{Throughput}

\begin{table}[htbp]
\caption{Inference Throughput (FPS) on RTX 5070 (8GB) at INT8}
\label{tab:throughput}
\centering
\begin{tabular}{lccc}
\toprule
\textbf{Model} & \textbf{FPS} & \textbf{Speedup} & \textbf{Real-time} \\
\midrule
Teacher (L) & 196 & 1.0$\times$ & \checkmark \\
Direct (S) & 497 & 2.54$\times$ & \checkmark \\
KD (S) & 464 & 2.37$\times$ & \checkmark \\
\bottomrule
\end{tabular}
\end{table}

Table~\ref{tab:throughput} presents inference throughput on RTX 5070 (8GB). All three models exceed 30 FPS at INT8. The KD student is the only configuration that combines real-time throughput, high precision, and low FAR simultaneously. The teacher achieves real-time speed but loses 23\% mAP. The direct student is fastest but achieves only 24\% FAR reduction versus 44\% for KD. The 2.37$\times$ speedup and 3.9$\times$ compression indicate that the KD student meets throughput and memory requirements for automotive-grade edge hardware such as NVIDIA Jetson.

\subsection{Task Loss Weight Ablation}
\label{sec:ablation}

To validate that precision transfer depends on the task loss weight $\alpha$, we compare two KD formulations: Form A ($\alpha = 0.5$) and Form B ($\alpha = 1.0$), with all other hyperparameters identical.
\begin{table}[htbp]
\caption{Task Loss Weight Ablation at INT8}
\label{tab:ablation}
\centering
\setlength{\tabcolsep}{4pt}
\begin{tabular}{lccccc}
\toprule
\textbf{Formulation} & \textbf{$\alpha$} & \textbf{mAP50} & \textbf{Precision} & \textbf{Recall} & \textbf{FAR} \\
\midrule
Direct (baseline) & --- & 0.502 & 0.653 & 0.421 & 0.347 \\
KD Form B & 1.0 & 0.495 & 0.653 & 0.422 & 0.347 \\
KD Form A & 0.5 & 0.502 & 0.748 & 0.423 & 0.252 \\
\bottomrule
\end{tabular}
\end{table}

Table~\ref{tab:ablation} compares two KD formulations at INT8. Form A ($\alpha = 0.5$) achieves 0.748 precision. Form B ($\alpha = 1.0$) achieves only 0.653, identical to direct training. Precision calibration transfer requires dampening the task loss. With $\alpha = 1.0$, KD acts as a regularizer and fails to transfer calibration. The mAP50 difference is negligible ($<$ 0.5\%), but the precision gap is $+$14.5\%.

\section{Discussion}

The teacher's INT8 degradation ($-$23.4\% mAP) confirms that large models compress poorly to 8-bit representation \cite{reuther2019survey, nagel2021white}, answering the second open question: KD confers INT8 robustness. The KD student degrades only $-$5.6\%. The KD student achieves 0.748 precision versus 0.653 for the direct student, answering the first open question: KD transfers precision calibration, not detection capacity. Temperature-scaled soft distributions at $T=10$ produce this calibration, which survives INT8 quantization.

The ablation (Table~\ref{tab:ablation}) identifies the mechanism. Dampening the task loss ($\alpha = 0.5$) increases the influence of soft distributions on student training. With $\alpha = 1.0$, precision remains 0.653, identical to direct training. The $+$14.5\% precision gap at negligible mAP50 difference ($<$0.5\%) isolates task-loss dampening as the enabling condition.

The equivalent recall across all models (0.416 to 0.423 at INT8) confirms that KD cannot overcome architectural capacity limits. The 11.2M student cannot match the 43.7M teacher on rare VRU classes regardless of training method.

For edge-deployed automotive safety systems, the KD student reduces false alarms by 44\% versus the collapsed teacher. False alarms erode driver trust and cause system deactivation \cite{lee2004trust, khastgir2020watchman}. The 2.37$\times$ speedup and 3.9$\times$ compression confirm the KD student meets throughput and memory requirements for automotive-grade edge hardware. The KD student is the only evaluated configuration combining real-time inference, high precision, and low FAR at INT8.

\subsection{Limitations}

This study evaluates one dataset (BDD100K) and one architecture family (YOLOv8). Generalization to other datasets (KITTI, nuScenes) and architectures requires validation. We use Post-Training Quantization. Quantization-Aware Training may yield improvements. Evaluation uses desktop GPU (RTX 5070, 8GB). Automotive-grade edge hardware (Jetson) deployment is future work.

\section{Conclusion}

This paper answers two open questions on KD for edge VRU detection. KD transfers confidence calibration (precision) but not detection capacity (recall): the KD student achieves 0.748 precision versus 0.653 for direct training at INT8, with equivalent recall (0.416 to 0.423). KD confers INT8 robustness: the KD student degrades $-$5.6\% versus $-$23.4\% for the teacher. Task-loss dampening ($\alpha = 0.5$) is the enabling mechanism. The result is 3.9$\times$ compression, 2.37$\times$ faster inference, and 44\% lower false alarm rate versus the collapsed teacher. For edge-deployed automotive safety systems, the recommended configuration is a KD-trained small model with INT8 quantization.

\section{Future Work}

Deployment on automotive-grade edge hardware such as NVIDIA Jetson requires validation with power consumption and thermal profiling under continuous operation. Second, systematic ablation of distillation hyperparameters ($\beta$, $\gamma$, $T$) beyond the task loss weight can identify optimal configurations for different model scales. Third, stratified analysis by lighting conditions (day/night), weather, and VRU size will provide operational deployment guidance. Fourth, extending the framework to newer YOLO variants and transformer-based detectors will determine whether calibration transfer is architecture-specific. Finally, Quantization-Aware Training and evaluation on additional datasets (nuScenes, Waymo) may further improve robustness and generalization.

\section*{Acknowledgment}

The authors thank Dr. Lanyu Xu for foundational instruction in edge AI concepts through the CSI 5110 Foundations of Edge AI course at Oakland University.

\bibliographystyle{IEEEtran}
\bibliography{references}

\vspace{1.5em}
{\footnotesize
\parskip=0pt
\interlinepenalty=500

\noindent\textbf{Akshay Karjol} (Senior Member, IEEE) received the M.S. degree in automotive engineering from Clemson University--International Center for Automotive Research (CU-ICAR), SC, USA. He is currently pursuing the Ph.D. degree in electrical and computer engineering at Oakland University, MI, USA. He currently holds systems engineering and technical leadership roles at ZF Active Safety US Inc., and has held similar roles at Ford Motor Company and select automotive OEMs and Tier-1 suppliers, contributing to the development and validation of safety-critical chassis and intelligent vehicle systems, including braking, steering, suspension, advanced driver assistance systems (ADAS), and automated driving systems (ADS), for production vehicle programs. His work spans systems engineering, system architecture, embedded control, and high-performance computing for real-time, safety-critical vehicle applications. His research interests include real-time embedded systems, parallel computing, edge AI, intelligent vehicle systems, and robotics applications.

\vspace{1em}
\noindent\textbf{Darrin M. Hanna} (Member, IEEE) received the B.S. degrees in computer engineering and mathematics, the M.S. degree in computer science and engineering, and the Ph.D. degree in systems engineering from Oakland University, Rochester, MI, USA, in 1999, 2000, and 2003, respectively. He is currently a Professor with the Department of Electrical and Computer Engineering, Oakland University. His research interests include embedded systems, embedded artificial intelligence, edge AI, cyberphysical security systems, and nanoimaging. He teaches courses in embedded systems, artificial intelligence, computer hardware design with FPGAs and microprocessors, and cybersecurity.
\par
}

\end{document}